\documentclass[sigconf]{acmart-arxiv}
\AtBeginDocument{%
  }

\usepackage{algorithm}
\usepackage{algorithmic}
\usepackage{balance}
\usepackage{colortbl}
\usepackage{multirow}

\copyrightyear{2026}
\acmYear{2026}
\setcopyright{cc}
\setcctype{by}
\acmDOI{10.1145/3767308.3835409}
\acmConference[MM '26] {Proceedings of the 35th ACM International Conference on Multimedia}{November 10--14, 2026}{Rio de Janeiro, Brazil.}
\acmBooktitle{Proceedings of the 35th ACM International Conference on Multimedia (MM '26), November 10--14, 2026, Rio de Janeiro, Brazil}
\acmISBN{979-8-4007-2213-4/2026/11}

\begin{document}

\title{SGPDFuse: Semantically-Guided Physics-Disentanglement General Multi-Modal Image Fusion}

\author{Haozhen Wei}
\orcid{0009-0000-9087-6425}
\affiliation{%
  \institution{Dalian University of Technology}
  \city{Dalian}
  \state{Liaoning}
  \country{China}
}
\email{wellwhz@163.com}

\author{Chengjun Jiang}
\orcid{0009-0009-1678-1650}
\affiliation{%
  \institution{Dalian University of Technology}
  \city{Dalian}
  \state{Liaoning}
  \country{China}
}
\email{20232241166@mail.dlut.edu.cn}

\author{Yutong Guo}
\orcid{0009-0009-0040-4754}
\affiliation{%
  \institution{Dalian University of Technology}
  \city{Dalian}
  \state{Liaoning}
  \country{China}
}
\email{yutongguo@mail.dlut.edu.cn}

\author{Xinrui Ju}
\orcid{0009-0004-9065-8689}
\affiliation{%
  \institution{City University of Hong Kong}
  \city{Hong Kong}
  \country{China}
}
\email{juxinrui1021@163.com}

\author{Xingyuan Li}
\orcid{0000-0001-9081-817X}
\affiliation{%
  \institution{Zhejiang University}
  \city{Hangzhou}
  \state{Zhejiang}
  \country{China}
}
\email{xingyuan\_lxy@163.com}

\author{Xiang Chen}
\orcid{0000-0002-0249-9664}
\affiliation{%
  \institution{Zhejiang University}
  \city{Hangzhou}
  \state{Zhejiang}
  \country{China}
}
\email{wasdnsxchen@gmail.com}

\author{Jinyuan Liu}
\orcid{0000-0003-2085-2676}
\authornote{Corresponding author.}
\affiliation{%
  \institution{Dalian University of Technology}
  \city{Dalian}
  \state{Liaoning}
  \country{China}
}
\email{atlantis918@hotmail.com}

\renewcommand{\shortauthors}{Haozhen Wei et al.}

\begin{abstract}
Multimodal image fusion (MMIF) aims to integrate complementary sensor data into a single representation that preserves \emph{intrinsic scene reality} while eliminating \emph{environmental interferences}. 
Most existing approaches rely on \emph{blind feature aggregation}, which excels at signal accumulation but fails to distinguish essential content from physical degradations.
We propose \textbf{SGPDFuse}, which bridges this gap by mapping inputs into a physics-disentangled structural representation via a \emph{Semantic-Physical Parametric Bridge} (SPPB) built on pretrained vision foundation models, utilizing the Intrinsic-Variation principle to decouple invariant scene attributes from transient environmental factors.
To guide this decomposition, we introduce a \emph{Semantic Alignment} mechanism: we explicitly anchor the fused representation to salient semantic features in the same foundation model feature space via cosine similarity to preserve critical targets, while enforcing physical texture fidelity through Gram-matrix regularization to strictly eliminate unnatural artifacts.
Extensive experiments demonstrate that SGPDFuse achieves state-of-the-art performance across infrared-visible, multi-focus, and multi-exposure benchmarks using a single architecture.
\end{abstract}



\begin{teaserfigure}
  \includegraphics[width=\textwidth]{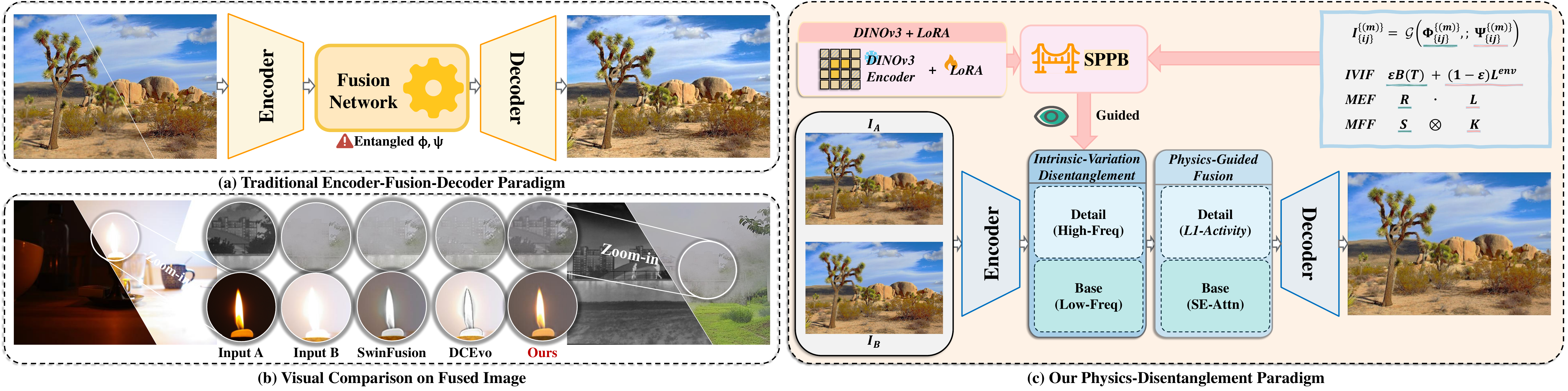}
  \caption{Paradigm shift in multi-modal fusion. (a) Traditional methods conflate \emph{intrinsic} scene content with \emph{variation} factors. (b) Our method recovers sharper detail. (c) SPPB predicts a variation map $\mathcal{W}$ from DINOv3 features, separating $\Phi$ from $\Psi$.}
  \label{fig:teaser}
\end{teaserfigure}


\maketitle

\section{Introduction}
\label{sec:intro}

Multimodal Image Fusion (MMIF) aims to integrate complementary information from different sensors into a single output that preserves essential scene content while suppressing sensor-specific degradations~\cite{li2024contourlet,li2026unifusion,li2026drfusion, li2026uncertainty, li2025difiisr}.
Taking infrared-visible fusion as a representative example, the fused result is expected to highlight thermally salient targets while retaining the rich textures and semantic readability of visible images, thereby alleviating challenges such as low illumination, noise contamination, and sensor-specific artifacts.
Meanwhile, fused images often serve as intermediate representations for downstream perception tasks, and their quality directly affects the robustness of subsequent systems such as object detection and segmentation.

Most existing deep fusion methods treat fusion as a \emph{signal aggregation} problem: given source images $\{I_A, I_B\}$, the objective is to learn a mapping that maximizes information preservation through weighted summation or feature concatenation.
We argue that this formulation suffers from an inherent limitation: it fundamentally conflates two distinct signal components---the \textbf{intrinsic} physical properties of the scene (true temperature, surface reflectance, sharp texture) and the \textbf{variation} factors introduced by environmental conditions (thermal reflections, illumination changes, optical blur).
This conflation manifests as persistent artifacts: ghost thermal targets on reflective surfaces, color distortions from exposure imbalance, or texture loss in defocused regions.
Consequently, existing methods struggle to simultaneously preserve essential content while suppressing environmental interference.

To address this, we propose a physics-disentanglement perspective by revisiting the fusion objective itself.
We draw inspiration from established physical imaging models: in infrared imaging, the observed radiance $L_{obs} = \epsilon B(T) + (1-\epsilon) L_{env}$ comprises thermal emission and environmental reflection~\cite{bao2023heat}; in photography, the Retinex model $I = R \cdot L$ separates reflectance from illumination~\cite{land1977retinex}; in optics, the defocus model $I = S \otimes K$ distinguishes sharp content from blur~\cite{subbarao1994depth,ruan2022learning}.
These diverse physical processes share a unified structure: $I = \mathcal{G}(\text{Intrinsic},\, \text{Variation})$.
This suggests that fusion should not blindly aggregate signals, but rather disentangle intrinsic content from variation factors before merging.

Motivated by this insight, we propose \textbf{SGPDFuse} (Semantically-Guided Physics-Disentanglement Multi-Modal Image Fusion), a unified framework that bridges high-level semantics with low-level physics.
Our key observation is that semantic features from vision foundation models---particularly DINOv3~\cite{simeoni2025dinov3}---encode material properties and object identities that are invariant to physical variations, making them ideal for predicting the variation parameters ($\epsilon$, illumination quality, focus measure) needed for disentanglement.
We instantiate this via a \emph{Semantic-Physical Parametric Bridge} (SPPB) that uses LoRA-adapted~\cite{hu2022lora} DINOv3 features to predict the variation confidence map $\mathcal{W}$, which drives Intrinsic-Variation disentanglement of both base (low-frequency) and detail (high-frequency) features.
To guide the decomposition, we further introduce a \emph{Semantic Alignment} mechanism that explicitly anchors the fused representation to salient semantic features via cosine similarity, while enforcing physical texture fidelity through Gram-matrix regularization to eliminate unnatural artifacts.
By coupling semantic understanding with physical priors, SGPDFuse enables adaptive physics-guided decomposition across infrared-visible, multi-exposure, and multi-focus fusion tasks within a single architecture.

\begin{figure}
    \centering
    \includegraphics[width=\linewidth]{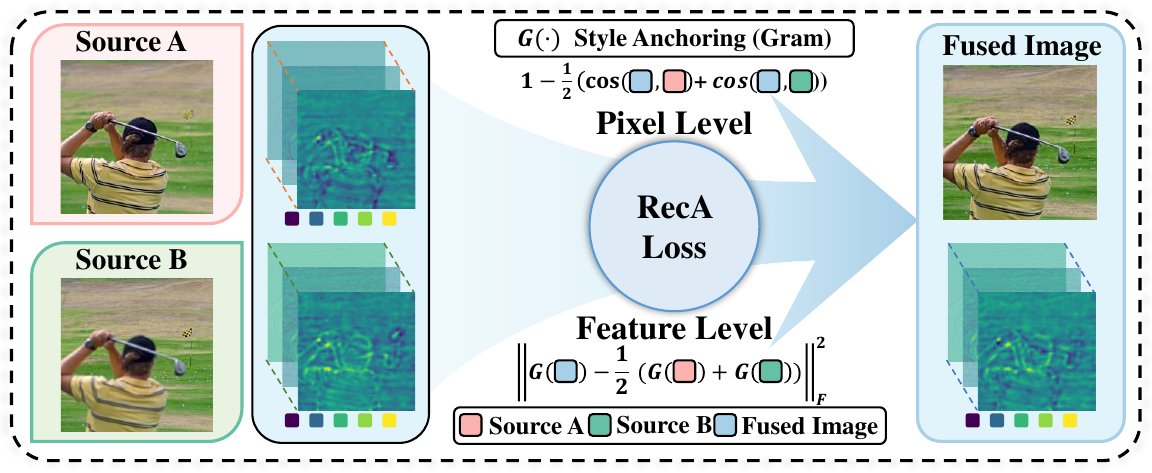}
    \caption{Illustration of our Reconstruction Alignment (RecA) loss for multi-modal image fusion.}
    \label{fig:RecA}
\end{figure}

Specifically, our contributions are:
\begin{itemize}
    \item We present a new perspective on MMIF by reformulating fusion from blind feature aggregation to physics-disentanglement under the unified Intrinsic-Variation principle, providing a framework to separate essential scene content from environmental interference across diverse fusion tasks.
    \item We design a Semantic-Physical Parametric Bridge (SPPB) that leverages DINOv3 features with LoRA adaptation to predict task-specific variation parameters, enabling unified physics-guided decomposition.
    \item We introduce a Semantic Alignment mechanism combining cosine-similarity-based salient feature preservation with Gram-matrix texture regularization, providing dual supervision that ensures both semantic fidelity and physical consistency of the fused output.
    \item Extensive experiments demonstrate that SGPDFuse achieves state-of-the-art performance across infrared-visible, multi-focus, and multi-exposure benchmarks using a single architecture, with consistent improvements in artifact suppression, detail preservation, and downstream task robustness.
\end{itemize}
\section{Related Work}
\label{sec:related}

\subsection{Task-Specific Multi-Modal Image Fusion}
Deep learning has driven remarkable progress in task-specific image fusion. Early CNN-based methods such as DenseFuse~\cite{li2018densefuse} and U2Fusion~\cite{xu2020u2fusion} established encoder-decoder paradigms for feature aggregation, while RFN-Nest~\cite{li2021rfn} introduced residual feedback mechanisms.
Subsequent methods further specialized their designs for distinct fusion scenarios: IVIF approaches~\cite{liu2022target,zhao2023metafusion,ma2019fusiongan} incorporate detection networks and adversarial training to preserve thermal saliency, MEF and MFF techniques~\cite{zhang2021benchmarking,zhang2021mff} leverage exposure-specific brightness decomposition and focus-map guidance, and PIAFusion~\cite{tang2022piafusion} exploits illumination-aware progressive training.
Transformer architectures have further extended modeling capacity, with SwinFusion~\cite{ma2022swinfusion} capturing cross-modal long-range dependencies and CDDFuse~\cite{zhao2023cddfuse} decomposing features into base and detail branches via transformer blocks.
More recently, Text-IF~\cite{yi2024text}, TC-MoA~\cite{zhu2024task}, RFfusion~\cite{wang2025efficient}, diffusion-based~\cite{zhao2023ddfm}, and task-driven~\cite{liu2024task} approaches further push the frontier via text conditioning, adapter mixtures, iterative denoising, and detection-aware strategies.

Despite their diversity, all these methods aggregate features from raw source signals, treating intrinsic scene content and extrinsic environmental factors as equally valid fusion targets.
This leads to artifacts when environmental interference is present, such as ghost thermal targets on reflective surfaces, texture loss in defocused regions, or unnatural halos under extreme exposure conditions.

\subsection{Physics-Informed and Disentanglement-Based Vision}
Explicit physical modeling has proven effective within individual imaging domains.
HADAR~\cite{bao2023heat} leverages the Stefan--Boltzmann radiation law $L = \epsilon B(T) + (1-\epsilon)L^{\text{env}}$ to disentangle thermal emission from environmental reflection in longwave infrared imaging.
Algorithm-unrolling methods~\cite{zhao2021efficient} also exploit physical models to drive network architecture design.
Retinex theory~\cite{land1977retinex} decomposes images as $I = R \cdot L$ (reflectance times illumination), spawning a family of deep enhancement methods~\cite{wei2018deep} for low-light and exposure correction.
Depth-from-Defocus~\cite{subbarao1994depth} and learning-based defocus deblurring~\cite{ruan2022learning} model the blurred image as $I = S \otimes K$, recovering the all-in-focus image $S$ by estimating the blur kernel $K$.
In representation learning, disentangled generative models~\cite{burgess2018understanding} aim to separate content from style, but these are trained for single-modality synthesis rather than cross-modal fusion.

These approaches share a common mathematical structure: the observed signal is a composite of intrinsic scene properties and extrinsic variation factors. 
While each of these approaches demonstrates the power of physical priors, they remain confined to single tasks, require domain-specific supervision (hyperspectral emissivity measurements, paired exposure data, depth ground truth).
Critically, the challenge of \emph{estimating} physical variation parameters from entangled multi-modal observations---without any explicit physical measurements---remains entirely unaddressed..

\textbf{Our Perspective.}
We observe that \emph{vision foundation models bridge this gap}: DINO~\cite{caron2021emerging,oquab2023dinov2,simeoni2025dinov3} semantic features encode material properties and structural priors invariant to the physical variations governing each imaging modality, enabling unified Intrinsic-Variation decomposition across diverse fusion tasks without explicit physical supervision.
\section{Intrinsic-Variation Structure in Multi-Modal Imaging}
\label{sec:preliminary}

A fundamental but underappreciated property of multi-modal imaging is that every observed modality is a \emph{composite} signal---it mixes the stable physical properties of the scene with transient, sensor-specific interference.
Existing fusion methods operate directly on these composite observations, implicitly assuming that sophisticated feature aggregation can filter out the undesired components.
We establish the theoretical foundation for SGPDFuse by formalizing a structural property that is universally present in multi-modal imaging yet has been overlooked in existing fusion methods.

\subsection{Latent-Factor Formulation of Image Formation}

Let $I^{(m)} \in \mathbb{R}^{H \times W}$ denote an observed image from modality $m$.
We model image formation at spatial location $(i,j)$ as a \emph{two-factor composition}:
\begin{equation}
     I^{(m)}_{ij} = \mathcal{G}\!\left(\Phi^{(m)}_{ij},\; \Psi^{(m)}_{ij}\right),
     \label{eq:formation}
\end{equation}
where $\Phi^{(m)}_{ij}$ is the \textbf{Intrinsic} component---the modality-invariant, scene-specific property that any high-fidelity fusion must preserve---$\Psi^{(m)}_{ij}$ is the \textbf{Variation} component---the environmental or sensor-induced factor that fusion must suppress---and $\mathcal{G}$ encodes the modality-specific physical mixing process.
Critically, both $\Phi$ and $\Psi$ are latent: the observation $I^{(m)}$ is an entangled composite in which neither factor is directly accessible.

\noindent\textbf{Why aggregation cannot resolve this entanglement.}
Consider the additive case $\mathcal{G}(\Phi,\Psi) = \Phi + \Psi$ and any spatially-adaptive fusion weight $\mathbf{w} \in [0,1]^{H \times W}$.
The aggregated output is:
\begin{equation}
     \hat{I} = \mathbf{w} \odot I^{(A)} + (\mathbf{1}{-}\mathbf{w}) \odot I^{(B)}
     = \Phi + \mathbf{w} \odot \Psi^{(A)} + (\mathbf{1}{-}\mathbf{w}) \odot \Psi^{(B)}.
     \label{eq:residual}
\end{equation}
Since physical variation fields are non-negative, the residual $\hat\Psi \geq 0$ is strictly positive at every pixel carrying variation---regardless of $\mathbf{w}$.
This \emph{irreducibility} holds for any fusion method that combines entangled signals before separating $\Phi$ from $\Psi$, implying that artifact-free fusion requires explicit $\Phi$ estimation and $\Psi$ suppression.

\subsection{Physical Instantiation Across Fusion Tasks}

Eq.~\eqref{eq:formation} is not an abstract construction---it is grounded in established physical imaging models across all three canonical fusion tasks.
In \textbf{IVIF}, $L = \epsilon B(T) + (1{-}\epsilon)L^{\text{env}}$~\cite{bao2023heat}: true thermal emission $\Phi = \epsilon B(T)$ mixes with environmental reflection $\Psi = (1{-}\epsilon)L^{\text{env}}$, producing ghost artifacts on low-emissivity surfaces.
In \textbf{MEF}, $I^{(k)} = R \cdot L^{(k)}$~\cite{land1977retinex}: reflectance $\Phi=R$ is entangled with exposure illumination $\Psi=L^{(k)}$, causing color distortions when fused without isolation.
In \textbf{MFF}, $I^{(k)} = S \otimes K(\sigma_k)$~\cite{subbarao1994depth}: the all-in-focus image $\Phi=S$ is convolved with blur kernel $\Psi=K(\sigma_k)$, producing mixed-gradient artifacts without kernel estimation.
Despite entirely different physical mechanisms, all three tasks instantiate the same abstract form Eq.~\eqref{eq:formation} with a common Intrinsic-Variation structure, making a \emph{task-unified} disentanglement architecture both principled and achievable.

\begin{figure*}[!ht]
    \centering
    \includegraphics[width=\textwidth]{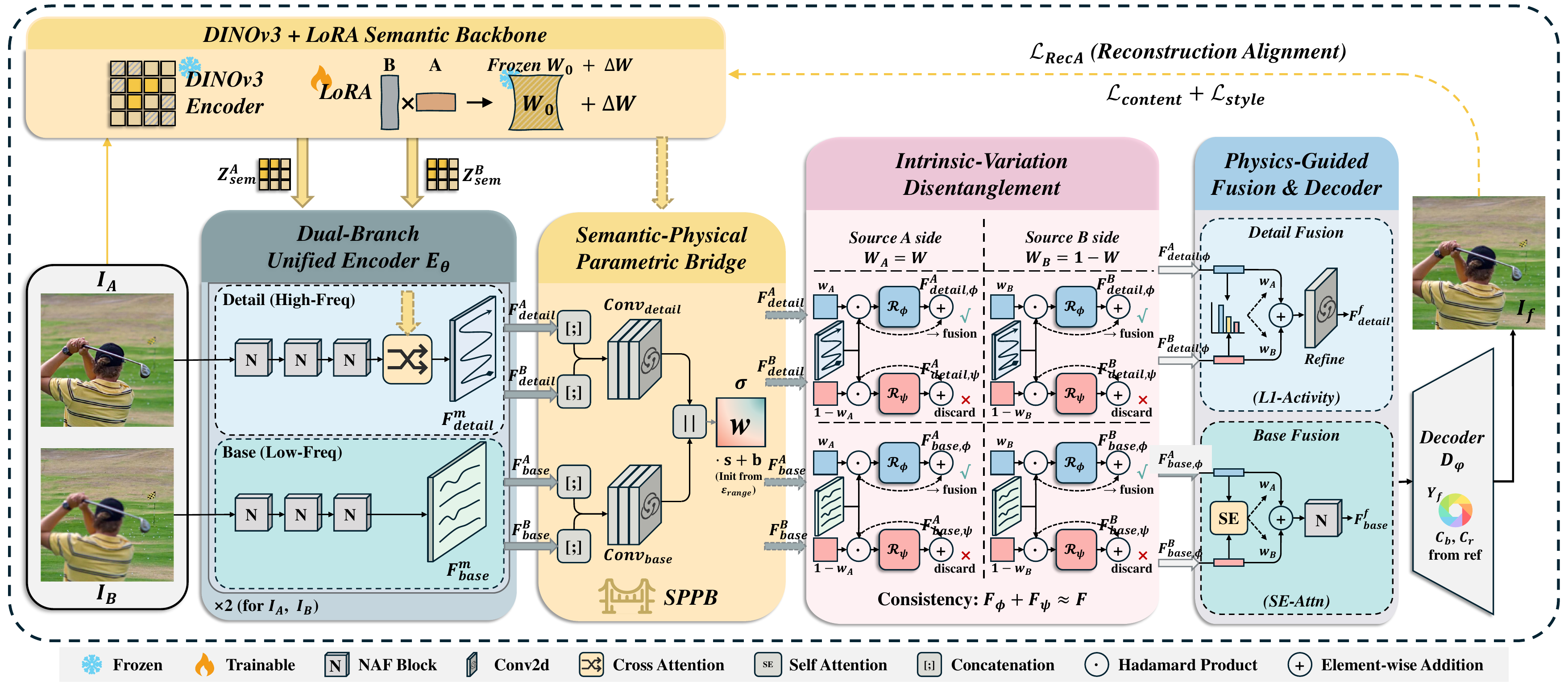}
    \caption{Overview of SGPDFuse. A dual-branch encoder extracts base and detail features from $I_A$, $I_B$; detail features are augmented with DINOv3-LoRA. The SPPB predicts a variation map $\mathcal{W}$, guiding disentanglement into intrinsic ($\Phi$) and variation ($\Psi$) components. Only $\Phi$ enters fusion; $\Psi$ is discarded. A semantic alignment loss $\mathcal{L}_{\text{sem}}$ supervises in DINO feature space.}
    \label{fig:pipeline}
\end{figure*}

\subsection{Scale-Aligned Factorization}

Eq.~\eqref{eq:residual} establishes that $\Psi$ must be estimated; the key question is \emph{how}.
A structural observation makes this tractable: across all three tasks, the Variation and Intrinsic components \emph{predominantly} occupy distinct frequency bands---$\Psi$ concentrates at low frequencies (smooth thermal gradients in IVIF, spatially slow illumination in MEF, low-pass blur attenuation in MFF), while $\Phi$ is most discriminable at high frequencies (edges, texture boundaries, sharp focus energy).
This \emph{approximate} spectral alignment is a statistical tendency grounded in physics, not an absolute partition.
It motivates a two-axis factorization that combines the Intrinsic-Variation decomposition with a Base-Detail signal decomposition:
\begin{equation}
     I^{(m)} = \mathcal{G}_{\text{b}}\!\left(\Phi_{\text{b}},\, \Psi_{\text{b}}\right)
     + \mathcal{G}_{\text{d}}\!\left(\Phi_{\text{d}},\, \Psi_{\text{d}}\right),
     \label{eq:twoaxis}
\end{equation}
where each frequency band independently undergoes Intrinsic-Variation disentanglement.

This yields a four-stage pipeline: (i)~decompose each source into base/detail; (ii)~estimate $\mathcal{W}$ via a semantically-grounded bridge; (iii)~disentangle $\Phi$ from $\Psi$ at each scale; (iv)~fuse only $\{\Phi_{\text{b}}, \Phi_{\text{d}}\}$, with variation eliminated by construction.
\section{Methodology}
\label{sec:method}

We present SGPDFuse, illustrated in Fig.~\ref{fig:pipeline}. Grounded in the two-axis factorization of Eq.~\eqref{eq:twoaxis}, the framework processes each source pair through a shared dual-branch encoder that extracts base and detail features, predicts the physical variation confidence map $\mathcal{W}$ via the Semantic-Physical Parametric Bridge (SPPB), disentangles intrinsic components $\Phi$ from variation components $\Psi$ at each frequency scale, and fuses only $\Phi$ through physics-guided aggregation. Training follows a two-stage curriculum: encoder--decoder reconstruction pre-training to establish a stable feature space, followed by end-to-end fusion fine-tuning with the full loss.

\subsection{Dual-Branch Feature Extraction}

Given two source images $I_A \in \mathbb{R}^{C_A \times H \times W}$ and $I_B \in \mathbb{R}^{C_B \times H \times W}$, a shared UnifiedEncoder $E_\theta$ maps each input into base and detail feature representations aligned with the factorization of Eq.~\eqref{eq:twoaxis}:
\begin{equation}
    \mathbf{F}^m_{\text{base}},\; \mathbf{F}^m_{\text{detail}} = E_\theta(I^m, Z^m_{\text{sem}}), \quad m \in \{A, B\},
\end{equation}
where $\mathbf{F}^m_{\text{base}} \in \mathbb{R}^{d \times \frac{H}{4} \times \frac{W}{4}}$ captures low-frequency energy where Variation operates, and $\mathbf{F}^m_{\text{detail}} \in \mathbb{R}^{d \times H \times W}$ captures high-frequency content where Intrinsic is most prominent, with $d=64$.

The encoder adopts NAFNet blocks~\cite{chen2022simple} for efficient local feature extraction. DINOv3 semantic tokens $Z^m_\text{sem}$ are injected into the \emph{detail branch only} via cross-attention after the second NAFNet block: detail features serve as queries, DINOv3 patch tokens as keys and values. This semantically grounds the high-frequency features while keeping the base branch purely convolutional. The semantic tokens are also passed separately to the SPPB for variation parameter estimation.

\subsection{Semantic-Physical Parametric Bridge}

The core innovation of SGPDFuse is the Semantic-Physical Parametric Bridge (SPPB), which translates high-level semantic understanding into low-level physical parameters---emissivity $\epsilon$ in IVIF, exposure optimality in MEF, in-focus probability in MFF.
These quantities are notoriously difficult to measure directly, yet DINOv3 patch tokens encode them implicitly: activations for metallic surfaces differ systematically from those for vegetation, reflecting distinct physical material properties.
The SPPB makes this correspondence explicit through a physics-grounded projection.

\paragraph{Semantic Feature Extraction.}
We employ DINOv3~\cite{simeoni2025dinov3} as the semantic backbone, adapted via LoRA~\cite{hu2022lora} to be task-aware.
For each source $I^{(m)}$, the adapted backbone extracts spatial semantic tokens $Z^{(m)} \in \mathbb{R}^{d_{\text{dino}} \times h \times w}$, encoding material identity and structural context invariant to the physical variations being estimated.

\paragraph{Material-Library Variation Estimation.}
Let $\mathcal{M} = \{(\varphi_k, \mathbf{e}_k)\}_{k=1}^{K}$ be a task-specific material library with measured physical values $\varphi_k$ (emissivity~\cite{bao2023heat} for IVIF; exposure optimality for MEF; focus probability for MFF) and DINOv3 prototype embeddings $\mathbf{e}_k$.
The SPPB retrieves a per-pixel anchor via soft nearest-neighbor regression  and combines it with encoder features:
\begin{align}
    \hat{\varphi}_{ij} & = \textstyle\sum_k \mathrm{softmax}_k\!\bigl({-}\|Z_{ij}{-}\mathbf{e}_k\|/\tau\bigr)\,\varphi_k, \label{eq:anchor} \\
    \mathcal{W}_{ij}   & = \sigma\!\bigl(f_\theta(\mathbf{F}^A, \mathbf{F}^B) + g(\hat{\varphi}_{ij})\bigr),
    \label{eq:sppb}
\end{align}
where $\hat{\varphi}_{ij}$ is the per-pixel material anchor retrieved from $\mathcal{M}$, $\tau$ is a temperature controlling retrieval sharpness, $f_\theta$ aggregates encoder features, and $g$ is a learnable scalar~projection.

\begin{algorithm}[!ht]
\caption{SGPDFuse Inference Pipeline}
\label{alg:sgpdfuse}
\begin{algorithmic}[1]
\REQUIRE Source images $I_A, I_B$; trained SGPDFuse model
\ENSURE Fused image $I_f$
\STATE \textbf{// Step 1: Semantic Feature Extraction}
\STATE $Z^A_{\text{sem}}, Z^B_{\text{sem}} \leftarrow \text{DINOv3}_{\text{LoRA}}(I_A),\; \text{DINOv3}_{\text{LoRA}}(I_B)$
\STATE \textbf{// Step 2: Dual-Branch Encoding}
\STATE $\mathbf{F}^m_{\text{base}}, \mathbf{F}^m_{\text{detail}} \leftarrow E_\theta(I^m, Z^m_{\text{sem}})$ for $m \in \{A,B\}$
\STATE \textbf{// Step 3: Physical Parameter Prediction (SPPB)}
\STATE $\mathcal{W} \leftarrow \varphi_{\min} + (\varphi_{\max}-\varphi_{\min})\cdot\sigma\!\bigl(f_\theta(\mathbf{F}^A,\mathbf{F}^B)\bigr)$
\STATE $\mathcal{W}_A \leftarrow \mathcal{W}$, \quad $\mathcal{W}_B \leftarrow 1 - \mathcal{W}$
\STATE \textbf{// Step 4: Intrinsic-Variation Disentanglement}
\FOR{$m \in \{A, B\}$, \; $\ell \in \{\text{base}, \text{detail}\}$}
    \STATE $\mathbf{F}^{m}_{\ell,\Phi} \leftarrow \mathcal{R}_\Phi(\mathcal{W}_m \odot \mathbf{F}^{m}_\ell) + \mathcal{W}_m \odot \mathbf{F}^{m}_\ell$
    \STATE $\mathbf{F}^{m}_{\ell,\Psi} \leftarrow \mathcal{R}_\Psi((1{-}\mathcal{W}_m) \odot \mathbf{F}^{m}_\ell) + (1{-}\mathcal{W}_m) \odot \mathbf{F}^{m}_\ell$
\ENDFOR
\STATE \textbf{// Step 5: Physics-Guided Fusion (Intrinsic Only)}
\STATE $\mathbf{F}^f_{\text{base}} \leftarrow \text{BaseFuse}(\mathbf{F}^A_{\text{base},\Phi},\; \mathbf{F}^B_{\text{base},\Phi},\; \mathcal{W})$
\STATE $\mathbf{F}^f_{\text{detail}} \leftarrow \text{DetailFuse}(\mathbf{F}^A_{\text{det},\Phi},\; \mathbf{F}^B_{\text{det},\Phi},\; \mathcal{W})$
\STATE \textbf{// Step 6: Color-Preserving Decoding}
\STATE $Y_f \leftarrow D_\phi(\mathbf{F}^f_{\text{base}},\; \mathbf{F}^f_{\text{detail}})$
\STATE $I_f \leftarrow \text{YCbCr2RGB}(Y_f,\; Cb_{\text{ref}},\; Cr_{\text{ref}})$
\RETURN $I_f$
\end{algorithmic}
\end{algorithm}

\subsection{Intrinsic-Variation Disentanglement}

Given the SPPB output $\mathcal{W}\in[0,1]^{H\times W}$, we perform a $\mathcal{W}$-weighted projection that separates $\Phi$ from $\Psi$ in feature space. Formally, $\mathcal{W}_m(i,j)$ approximates the intrinsic fraction $\|\Phi^m\|/(\|\Phi^m\|+\|\Psi^m\|)$ at each location, so $\mathcal{W}_m\odot\mathbf{F}^m$ retains intrinsic content while suppressing variation. Learnable residual networks $\mathcal{R}_\Phi, \mathcal{R}_\Psi$ correct for non-orthogonality between the two components:
\begin{align}
    \mathbf{F}^m_\Phi & = \mathcal{W}_m \odot \mathbf{F}^m + \mathcal{R}_\Phi(\mathcal{W}_m \odot \mathbf{F}^m), \label{eq:disent_int}             \\
    \mathbf{F}^m_\Psi & = (1{-}\mathcal{W}_m) \odot \mathbf{F}^m + \mathcal{R}_\Psi((1{-}\mathcal{W}_m) \odot \mathbf{F}^m). \label{eq:disent_var}
\end{align}
A structural consistency constraint $\mathbf{F}^m_\Phi + \mathbf{F}^m_\Psi \approx \mathbf{F}^m$ ensures that no information is lost or duplicated during decomposition, regularizing $\mathcal{R}_\Phi$ and $\mathcal{R}_\Psi$ to act as complementary correctors rather than independent feature extractors.
This constraint is explicitly enforced via a reconstruction penalty $\mathcal{L}_\text{cons} = \|\mathbf{F}^m_\Phi + \mathbf{F}^m_\Psi - \mathbf{F}^m\|_2^2$, included in $\mathcal{L}_\text{decomp}$ (Eq.~\eqref{eq:decomp}) alongside the cross-correlation terms, which together prevent the residual networks from bypassing the disentanglement logic.

\paragraph{Complementary Source Assignment.}
We assign $\mathcal{W}_A = \mathcal{W}$ and $\mathcal{W}_B = 1{-}\mathcal{W}$, encoding a physically motivated complementarity derived from the image formation models.
In IVIF, the radiation law $L = \epsilon B(T) + (1{-}\epsilon)L^\text{env}$ directly induces this duality: at a pixel where emissivity $\epsilon$ is high, the IR channel is dominated by genuine emission ($\Phi$ for source $A$), while the visible channel captures mostly reflected illumination ($\Psi$ for source $B$); at low-$\epsilon$ surfaces, the roles reverse.
In MEF and MFF, the complementarity follows from energy conservation across exposure pairs and from focus-defocus duality, respectively.
We acknowledge this is an approximation; at pixels where both modalities carry distinct intrinsic information, the learnable residual correctors $\mathcal{R}_\Phi, \mathcal{R}_\Psi$ adaptively refine the decomposition where the hard assignment is insufficient. At both base and detail scales, $\Psi$ components are discarded before fusion, eliminating variation by construction.

\subsection{Physics-Guided Fusion}

Only the intrinsic components $\{\mathbf{F}^m_{\Phi}\}$ enter the fusion stage, by construction free of variation residual. The fusion strategy differs by frequency scale, reflecting the distinct physical nature of intrinsic content at each level.

The \textbf{base layer} captures low-frequency structure where complementary information should be weighted by relative physical confidence. We apply squeeze-and-excitation (SE) channel attention biased by $\mathcal{W}$, so that the higher-confidence source governs base-level fusion:
\begin{align}
    [w_A,w_B]                  & = \text{Softmax}\!\bigl(\text{SE}([\mathbf{F}^A_{\text{base},\Phi};\,\mathbf{F}^B_{\text{base},\Phi}])\bigr), \label{eq:basefuse} \\
    \mathbf{F}^f_{\text{base}} & = \text{NAF}\!\bigl(\textstyle\sum_m w_m\odot\mathbf{F}^m_{\text{base},\Phi}\bigr). \notag
\end{align}

The \textbf{detail layer} carries high-frequency content where the primary criterion is local sharpness. We adopt $\ell_1$-norm activity weighting as a proxy for signal energy, selecting the sharper modality at each location:
\begin{equation}
    w_m^{\text{d}}=\frac{\|\mathbf{F}^m_{\text{detail},\Phi}\|_1}{\sum_m\|\mathbf{F}^m_{\text{detail},\Phi}\|_1},\quad
    \mathbf{F}^f_{\text{detail}}=\text{Refine}\!\bigl(\textstyle\sum_m w_m^{\text{d}}\odot\mathbf{F}^m_{\text{detail},\Phi}\bigr).
    \label{eq:detailfuse}
\end{equation}
This asymmetric design---confidence weighting at base level, activity-based selection at detail level---directly follows from the Intrinsic-Variation factorization: with variation removed, fusion reduces to combining clean intrinsic signals according to their frequency-domain properties.

\begin{figure*}[!ht]
    \centering
    \includegraphics[width=\textwidth]{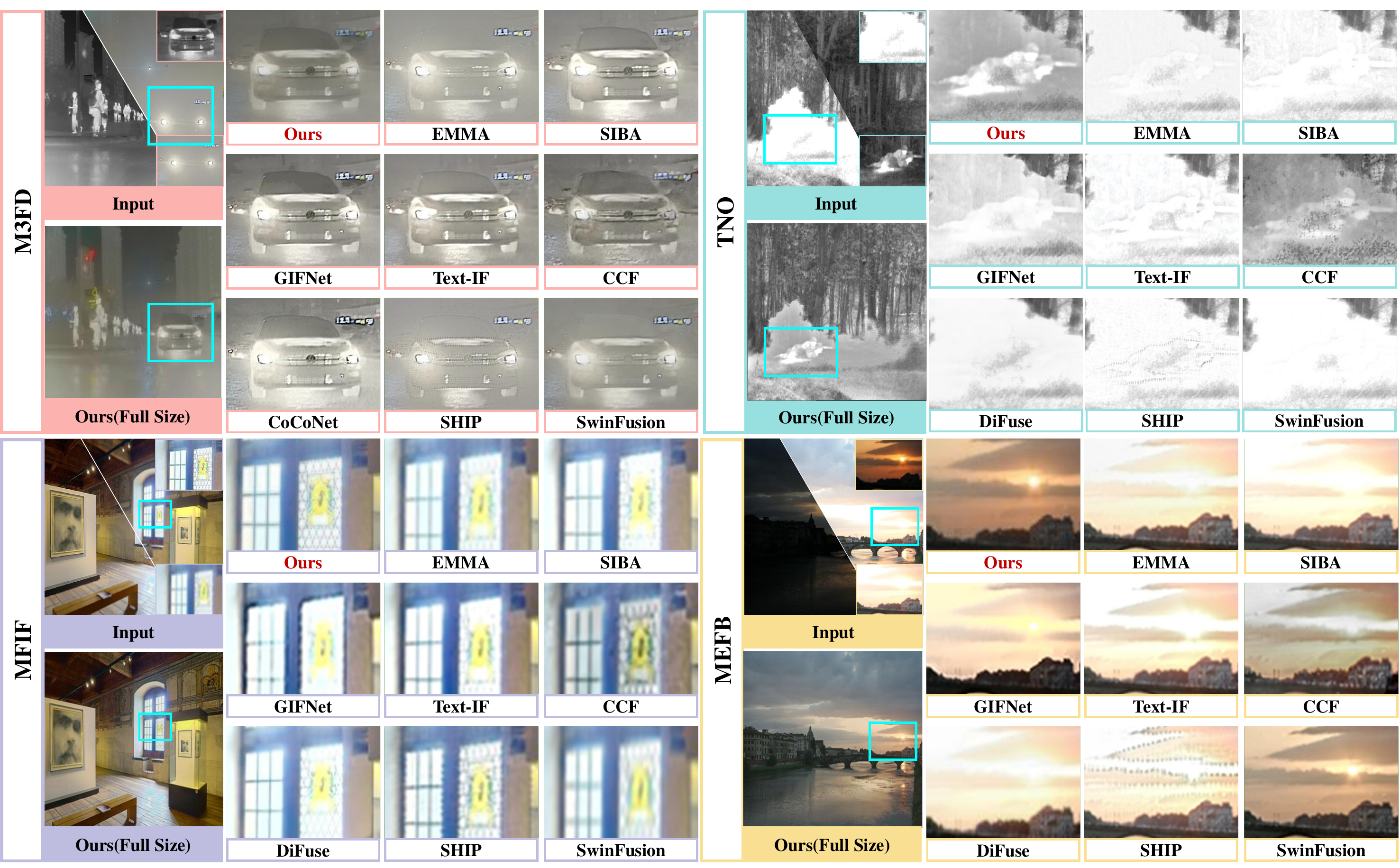}
    \caption{Qualitative comparison of fusion quality with other state-of-the-art methods on diverse multimodal fusion tasks. Magnified views demonstrate that SGPDFuse consistently achieves superior visual quality with enhanced detail preservation.}
    \label{fig:comparison}
\end{figure*}

\subsection{Loss Function}

Stage~1 pre-trains the encoder--decoder via $\mathcal{L}_{\text{rec}}$:
\begin{equation}
    \mathcal{L}_{\text{rec}} = \sum_{m\in\{A,B\}} \bigl(\|\hat{I}_m - I_m\|_1 + 0.5\,\mathcal{L}_{\text{SSIM}}(\hat{I}_m, I_m)\bigr).
    \label{eq:rec}
\end{equation}

Stage~2 trains the full pipeline end-to-end with:
\begin{equation}
    \mathcal{L}_{\text{total}} = \mathcal{L}_{\text{IG}} + \lambda_{\text{decomp}}\mathcal{L}_{\text{decomp}} + \lambda_{\text{RecA}}\mathcal{L}_{\text{RecA}} + \lambda_{\text{tv}}\mathcal{L}_{\text{tv}},
    \label{eq:total}
\end{equation}
where $\mathcal{L}_{\text{IG}}$, $\mathcal{L}_{\text{decomp}}$, and $\mathcal{L}_{\text{tv}}$ are formulated as:
\begin{equation}
    \mathcal{L}_{\text{IG}} = \|I_f - \max(I_A, I_B)\|_1 + \lambda_g \|\nabla I_f - \max(\nabla I_A, \nabla I_B)\|_1,
    \label{eq:IG}
\end{equation}
\begin{equation}
    \mathcal{L}_{\text{decomp}} = \frac{\operatorname{cc}(\mathbf{F}^A_{\text{d}},\, \mathbf{F}^B_{\text{d}})^2}{1 + \operatorname{cc}(\mathbf{F}^A_{\text{b}},\, \mathbf{F}^B_{\text{b}})} + \sum_{m}\|\mathbf{F}^m_\Phi {+} \mathbf{F}^m_\Psi {-} \mathbf{F}^m\|_2^2,
    \label{eq:decomp}
\end{equation}
\begin{equation}
    \mathcal{L}_{\text{tv}} = \sum_{i,j}\bigl(|\mathcal{W}_{i+1,j}-\mathcal{W}_{i,j}|+|\mathcal{W}_{i,j+1}-\mathcal{W}_{i,j}|\bigr),
    \label{eq:tv}
\end{equation}
where $I_f$ is the fused output, $\nabla$ denotes spatial gradients computed by finite differences, $\operatorname{cc}(\cdot,\cdot)$ is the Pearson cross-correlation with subscripts b/d denoting base/detail, and $\mathcal{W}$ is the variation confidence map from the SPPB.
The intensity-gradient loss enforces pixel and gradient fidelity to the source maximum; the decomposition loss encourages correlated base and complementary detail features while constraining $\mathcal{R}_\Phi, \mathcal{R}_\Psi$ as correctors; and the total-variation regularizer promotes spatial smoothness in $\mathcal{W}$.

\paragraph{Reconstruction Alignment Loss.}
Illustrated in Fig.~\ref{fig:RecA}, $\mathcal{L}_{\text{RecA}}$ anchors the fused output in DINOv3 feature space via cosine content alignment and Gram-matrix style regularization. Let $Z_f, Z_A, Z_B$ denote the DINOv3 features of the fused and source images:
\begin{align}
    \mathcal{L}_{\text{RecA}}    & = \alpha\,\mathcal{L}_{\text{content}} + \beta\,\mathcal{L}_{\text{style}}, \label{eq:RecA} \\
    \mathcal{L}_{\text{content}} & = 1 - \tfrac{1}{2}\bigl(\cos(Z_f,Z_A)+\cos(Z_f,Z_B)\bigr), \label{eq:content}               \\
    \mathcal{L}_{\text{style}}   & = \bigl\|G(Z_f)-\tfrac{1}{2}\bigl(G(Z_A)+G(Z_B)\bigr)\bigr\|_F^2, \label{eq:style}
\end{align}
where $\cos(\cdot,\cdot)$ is cosine similarity and $G(\cdot)$ is the Gram matrix.
\section{Experiments}
\label{sec:exp}

\begin{table*}[!ht]
\centering
\setlength{\tabcolsep}{4pt}
\renewcommand{\arraystretch}{1.25}
\caption{Quantitative comparisons of fusion metrics with other SOTA fusion methods on $\text{M}^3\text{FD}$, TNO \& RoadScene, MFIF and MEFB datasets.\textbf{Boldface} denotes the best while \underline{underline} denotes the second best results.}
\resizebox{\textwidth}{!}{
\setlength{\tabcolsep}{3pt}
\renewcommand{\arraystretch}{1.25}
\begin{tabular}{l|cccc|cccc|cccc|cccc}
\noalign{\hrule height 1.2pt}
\multirow{2}{*}{\textbf{Methods}} 
& \multicolumn{4}{c|}{\textbf{$\text{M}^3\text{FD}$ Dataset}} 
& \multicolumn{4}{c|}{\textbf{T \& R Dataset}} 
& \multicolumn{4}{c|}{\textbf{MFIF Dataset}} 
& \multicolumn{4}{c}{\textbf{MEFB Dataset}} \\
\cline{2-17}
 & CC\(\uparrow\) & PSNR\(\uparrow\) & TE\(\uparrow\) & SSIM\(\uparrow\) 
 & CC\(\uparrow\) & PSNR\(\uparrow\) & TE\(\uparrow\) & SSIM\(\uparrow\) 
 & CC\(\uparrow\) & PSNR\(\uparrow\) & TE\(\uparrow\) & SSIM\(\uparrow\) 
 & CC\(\uparrow\) & PSNR\(\uparrow\) & TE\(\uparrow\) & SSIM\(\uparrow\) \\
\hline
SwinFusion \cite{ma2022swinfusion}
& 0.52 & 62.65 & \underline{7.10} & 0.44 
& 0.61 & 61.41 & \underline{7.16} & 0.54 
& \underline{0.97} & \underline{75.73} & 7.29 & \underline{0.91} 
& 0.89 & \textbf{59.05} & 4.90 & \underline{0.80} \\
TC-MoA \cite{zhu2024task}
& 0.48 & 61.92 & 6.93 & 0.49 
& 0.58 & 62.89 & 6.86 & 0.52 
& 0.97 & 74.87 & 7.29 & 0.90 
& 0.86 & 58.92 & 4.88 & 0.80 \\
GIFNet \cite{cheng2025one}
& \textbf{0.58} & \underline{63.36} & 6.74 & 0.51 
& \textbf{0.64} & 63.06 & 6.87 & 0.54 
& 0.95 & 68.17 & 7.14 & 0.83 
& \underline{0.89} & 58.12 & 4.90 & 0.56 \\
CCF \cite{cao2024conditional}
& 0.53 & 63.27 & 6.89 & \underline{0.69} 
& 0.57 & \underline{63.43} & 6.94 & \underline{0.59} 
& 0.90 & 68.81 & 7.13 & 0.73 
& 0.85 & 58.65 & 4.55 & 0.50 \\
EMMA \cite{zhao2024equivariant}
& 0.51 & 61.76 & 6.83 & 0.51 
& 0.61 & 62.21 & 6.73 & 0.53 
& 0.97 & 70.69 & 7.14 & 0.88 
& 0.85 & 56.64 & 5.23 & 0.57 \\
DCEvo \cite{liu2025dcevo}
& 0.50 & 61.35 & 6.90 & 0.51 
& 0.59 & 61.98 & 7.00 & 0.53 
& 0.97 & 71.92 & 7.16 & 0.89 
& 0.84 & 56.32 & 5.23 & 0.58 \\
SIBA \cite{wang2025source}
& 0.53 & 61.55 & 6.83 & 0.52 
& 0.60 & 62.35 & 6.91 & 0.52 
& 0.97 & 71.51 & 7.24 & 0.89 
& 0.85 & 56.30 & \underline{5.67} & 0.65 \\
Text-DiFuse \cite{zhang2024text}
& 0.41 & 60.02 & 6.70 & 0.23 
& 0.50 & 60.51 & 7.05 & 0.26 
& 0.88 & 67.04 & 7.13 & 0.68 
& 0.80 & 56.86 & 4.95 & 0.40 \\
SHIP \cite{zheng2024probing}
& 0.47 & 61.24 & 6.93 & 0.49 
& 0.57 & 61.97 & 7.03 & 0.50 
& 0.92 & 69.19 & \underline{7.30} & 0.84 
& 0.89 & 56.19 & 5.47 & 0.60 \\
Text-IF \cite{yi2024text}
& 0.50 & 62.10 & 6.82 & 0.50 
& 0.60 & 62.77 & 6.80 & 0.51 
& 0.97 & 71.45 & 7.15 & 0.86 
& 0.86 & 56.39 & 5.44 & 0.76 \\
\hline
\rowcolor{gray!20}
\textbf{Ours}
& \underline{0.56} & \textbf{63.57} & \textbf{7.16} & \textbf{0.76} 
& \textbf{0.64} & \textbf{64.05} & \textbf{7.23} & \textbf{0.68} 
& \textbf{0.98} & \textbf{77.17} & \textbf{7.35} & \textbf{0.96} 
& \textbf{0.90} & \underline{58.94} & \textbf{5.88} & \textbf{0.82} \\
\noalign{\hrule height 1.2pt}
\end{tabular}
}
\label{tab:fusion}
\end{table*}

\begin{table*}[!ht]
\centering
\small
\setlength{\tabcolsep}{4pt}
\renewcommand{\arraystretch}{1.25}
\caption{Quantitative comparisons of downstream task performance with other existing fusion methods for object detection on $\text{M}^3\text{FD}$ and semantic segmentation on FMB dataset. \textbf{Boldface} denotes the best while \underline{underline} denotes the second best results.}
\setlength{\tabcolsep}{3pt}
\renewcommand{\arraystretch}{1.25}
\resizebox{\linewidth}{!}{
\begin{tabular}{l|ccccccc|ccccccc}
\noalign{\hrule height 1.2pt}
\multirow{2}{*}{\textbf{Methods}} & \multicolumn{7}{c|}{\textbf{$\text{M}^3\text{FD}$ Dataset}} & \multicolumn{7}{c}{\textbf{FMB Dataset}} \\
\cline{2-15} 
 & People & Car & Bus & Light & Moto & Trunk & mAP & Car & Trunk & T-Light & T-Sign & Motor & Pole & mIoU \\
\hline
SwinFusion \cite{ma2022swinfusion}
& 0.283 & 0.495 & 0.624 & 0.107 & 0.178 & 0.400 & 0.348 
& 0.821 & 0.505 & 0.268 & 0.568 & 0.684 & 0.403 & 0.665 \\
TC-MoA \cite{zhu2024task}
& 0.288 & \underline{0.501} & 0.624 & \textbf{0.122} & 0.155 & 0.405 & 0.349 
& 0.830 & 0.490 & \textbf{0.366} & \underline{0.703} & 0.644 & 0.415 & 0.687 \\
GIFNet \cite{cheng2025one}
& 0.280 & 0.496 & 0.610 & 0.106 & 0.141 & \underline{0.429} & 0.344 
& \textbf{0.834} & \underline{0.579} & 0.327 & 0.687 & 0.690 & 0.419 & \underline{0.695} \\
CCF \cite{cao2024conditional}
& 0.257 & 0.481 & 0.608 & 0.105 & 0.106 & 0.405 & 0.327 
& 0.813 & 0.464 & 0.317 & 0.694 & 0.663 & 0.416 & 0.678 \\
EMMA \cite{zhao2024equivariant}
& 0.276 & 0.496 & 0.624 & 0.118 & 0.147 & 0.398 & 0.343 
& \textbf{0.834} & 0.549 & 0.343 & 0.668 & 0.676 & 0.417 & 0.691 \\
DCEvo \cite{liu2025dcevo}
& 0.289 & 0.498 & 0.618 & 0.117 & 0.147 & 0.398 & 0.344 
& 0.829 & 0.518 & 0.349 & 0.654 & 0.686 & 0.421 & 0.687 \\
SIBA \cite{wang2025source}
& \underline{0.291} & 0.496 & \underline{0.633} & 0.109 & 0.103 & 0.399 & 0.338 
& 0.831 & 0.484 & 0.337 & 0.666 & 0.650 & 0.420 & 0.680 \\
Text-DiFuse \cite{zhang2024text}
& 0.263 & 0.480 & 0.602 & 0.082 & 0.127 & 0.423 & 0.329 
& 0.822 & 0.521 & 0.316 & 0.687 & 0.682 & 0.413 & 0.684 \\
SHIP \cite{zheng2024probing}
& 0.285 & 0.491 & \textbf{0.634} & 0.117 & \underline{0.185} & 0.402 & 0.352 
& 0.825 & 0.479 & 0.357 & 0.676 & \underline{0.696} & \textbf{0.431} & 0.688 \\
Text-IF \cite{yi2024text}
& 0.287 & \underline{0.501} & 0.624 & \underline{0.121} & 0.170 & 0.416 & \underline{0.353} 
& 0.831 & 0.503 & 0.318 & 0.680 & 0.675 & 0.424 & 0.684 \\
\hline
\rowcolor{gray!20}
\textbf{Ours}
& \textbf{0.297} & \textbf{0.532} & 0.564 & 0.102 & \textbf{0.205} & \textbf{0.440} & \textbf{0.357} 
& 0.833 & \textbf{0.601} & \underline{0.358} & \textbf{0.703} & \textbf{0.701} & \textbf{0.431} & \textbf{0.698} \\
\noalign{\hrule height 1.2pt}
\end{tabular}
}
\label{table:downstream}
\end{table*}

\subsection{Experimental Setup}

\paragraph{Experiment datasets}
To evaluate the performance of our method, we conduct experiments on three representative image fusion tasks: infrared and visible image fusion (IVIF), multi-exposure image fusion (MEF) and multi-focus image fusion (MFF). 
For IVIF, we use three widely adopted benchmarks: \(\text{M}^3\text{FD}\)~\cite{liu2022target}, RoadScene~\cite{xu2020fusiondn}, and TNO~\cite{toet2017tno}.
For MEF and MFF, we utilize the MEFB~\cite{zhang2021benchmarking} and MFIF~\cite{zhang2021deep} datasets.
For downstream evaluation on IVIF, we use $\text{M}^3\text{FD}$~\cite{liu2022target} for object detection and FMB~\cite{liu2023multi} for semantic segmentation.

\paragraph{Implementation Details}
Input images are resized to the nearest $16{\times}$ multiple of their original resolution; test images are evaluated at full resolution without any post-processing.
The shared encoder uses $d{=}64$ feature channels with $N{=}4$ NAFNet blocks; the SPPB uses 32 hidden channels per branch.
We adopt DINOv3 ViT-S/16 ($d_{\text{dino}}{=}384$) fine-tuned via LoRA with rank $r{=}32$ and $\alpha{=}64$.
Both stages use AdamW (weight decay $1{\times}10^{-4}$) with cosine annealing ($\eta_{\min}{=}10^{-6}$) and batch size~2.
\textbf{Stage~1} (reconstruction pre-training): 50 epochs, lr${=}1{\times}10^{-4}$, $\ell_1$+SSIM loss.
\textbf{Stage~2} (end-to-end fusion): 100 epochs, lr${=}5{\times}10^{-5}$; loss coefficients $\lambda_{\text{decomp}}{=}2.0$, $\lambda_{\text{tv}}{=}5.0$, $\lambda_g{=}10.0$, $\lambda_{\text{RecA}}{=}0.1$.
For IVIF, the SPPB scale and bias are initialized from the emissivity database derived from HADAR~\cite{bao2023heat}.
All experiments are conducted on 8 NVIDIA RTX3090 (24GB) GPUs using PyTorch.

\paragraph{Evaluation Metrics}
We adopt four complementary metrics from IVIF-ZOO~\cite{liu2024infrared}: Correlation Coefficient (CC), Peak Signal-to-Noise Ratio (PSNR), Transfer Entropy (TE), and Multi-Scale Structural Similarity (MS-SSIM, reported as SSIM in Table~\ref{tab:fusion}).
For downstream tasks, we use YOLOv8s~\cite{varghese2024yolov8} with mAP$_{50:95}$ for object detection and SegFormer-B1~\cite{xie2021segformer} with mIoU for semantic segmentation. For downstream IVIF application, input images are resized to $256\times256$.

\begin{figure*}
    \centering
    \includegraphics[width=\linewidth]{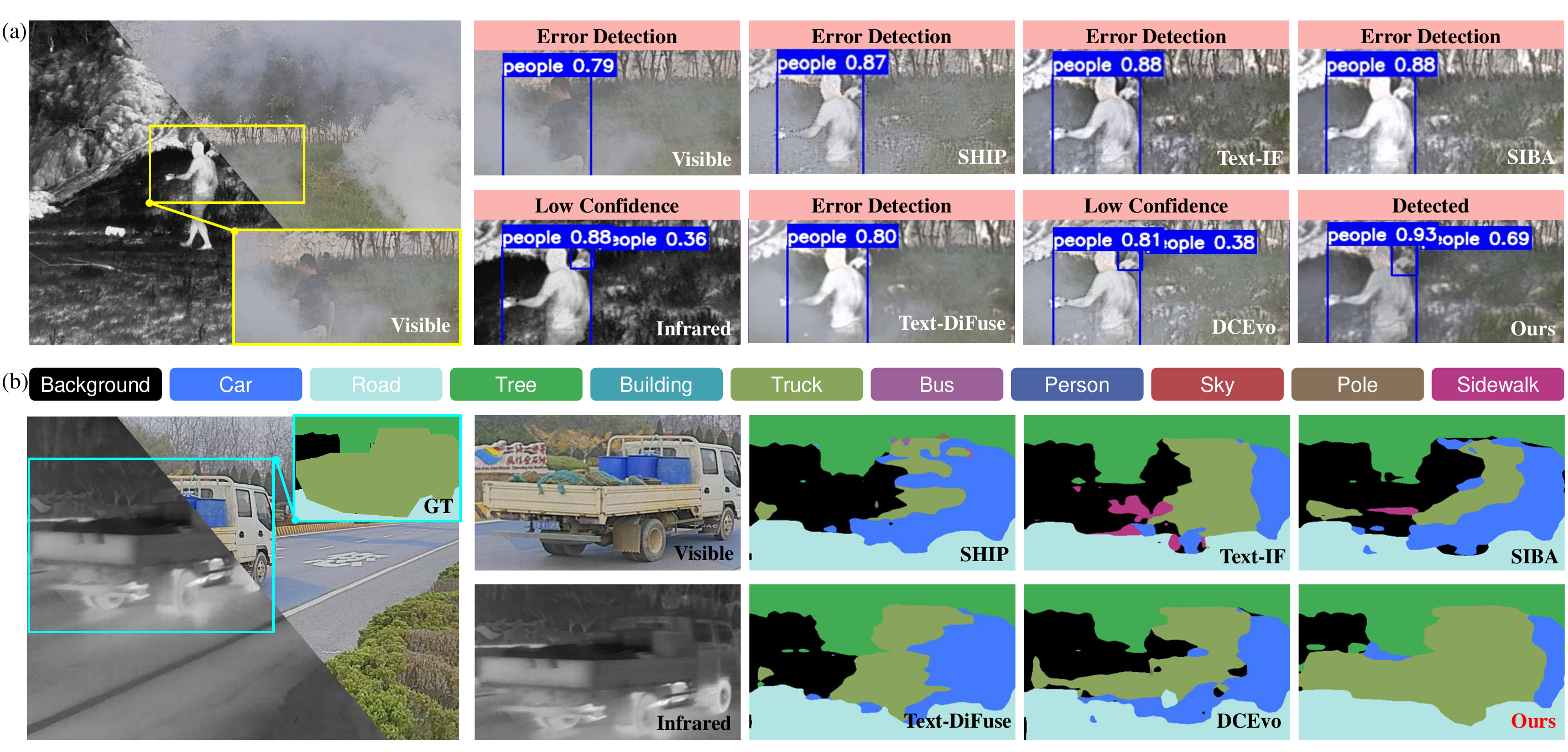}
    \caption{(a)Qualitative comparisons of object detection performance in smoke scene. (b)Qualitative comparisons of semantic segmentation performance in nighttime scene.}
    \label{fig:downstream_comparison}
\end{figure*}

\subsection{Experiments on Multi-Task}

\subsubsection{Qualitative Comparison}
Figure~\ref{fig:comparison} presents visual comparisons across all four benchmarks.
For IVIF, our method produces sharper vehicle outlines and headlight structures on $\text{M}^3\text{FD}$ and retains fine-grained branch and ground textures on TNO, where competing methods exhibit blurring or halo artifacts due to unresolved thermal emissivity variation.
For MFF, our results achieve the clearest focus transitions, with well-defined stained glass patterns that competitors render with ringing or local blur.
For MEF, our method jointly recovers detail from over-exposed and under-exposed regions, yielding naturally balanced luminance free of tone-mapping artifacts.
These consistent gains across tasks validate that SPPB-guided intrinsic–variation disentanglement effectively suppresses modality-specific artifacts.

\subsubsection{Quantitative Comparison}
Table~\ref{tab:fusion} presents quantitative comparisons across four benchmarks spanning IVIF ($\text{M}^3\text{FD}$, T\&R), MFF (MFIF), and MEF (MEFB).
Our method achieves the best or second-best on all datasets and metrics, with particularly decisive gains in MS-SSIM: +0.07 on $\text{M}^3\text{FD}$, +0.09 on T\&R, +0.05 on MFIF, and +0.02 on MEFB over the respective runners-up.
Consistent TE and PSNR gains across all benchmarks confirm that the physics-calibrated confidence maps enable complete intrinsic transfer while suppressing cross-modal noise, with improvements spanning three tasks under distinct physical priors---validating the generalizability of the architecture.

\subsection{Downstream Application on IVIF}
To further validate the practical utility of our fused images, we evaluate downstream performance on object detection and semantic segmentation.
As shown in Table~\ref{table:downstream}, our method achieves the highest mAP$_{50:95}$ on $\text{M}^3\text{FD}$ and the best mIoU on FMB among all competing methods.
The gains on thermally distinctive categories (Car, Motorcycle, Trunk) demonstrate that SPPB-calibrated intrinsic fusion preserves object signatures more faithfully, while mIoU improvements on FMB confirm that Intrinsic-Variation disentanglement maintains semantic boundaries essential for recognition.

\subsection{Ablation Studies}

\begin{table}[t]
\centering
\setlength{\tabcolsep}{5pt}
\renewcommand{\arraystretch}{1.15}
\caption{Ablation study on $\text{M}^3\text{FD}$. \textbf{Bold} denotes the best result.}
\resizebox{\linewidth}{!}{
\begin{tabular}{l|cccc}
\noalign{\hrule height 1.2pt}
\textbf{Variant} & CC\(\uparrow\) & PSNR\(\uparrow\) & TE\(\uparrow\) & SSIM\(\uparrow\) \\
\noalign{\hrule height 0.8pt}
\multicolumn{5}{l}{\textit{Study on SPPB}} \\
\hline
\textit{(A)} Full model & \textbf{0.56} & \textbf{63.57} & \textbf{7.16} & \textbf{0.76} \\
\textit{(B)} w/o SPPB ($\mathcal{W}=0.5$) & 0.52 & 61.38 & 7.10 & 0.72 \\
\textit{(C)} w/o IV Disentanglement & 0.53 & 61.45 & 7.10 & 0.73 \\
\textit{(D)} w/o LoRA (frozen DINOv3) & 0.50 & 60.22 & 6.47 & 0.69 \\
\noalign{\hrule height 0.8pt}
\multicolumn{5}{l}{\textit{Study on RecA Loss}} \\
\hline
\textit{(A)} Full model & \textbf{0.56} & \textbf{63.57} & \textbf{7.16} & \textbf{0.76} \\
\textit{(E)} w/o $\mathcal{L}_{\text{RecA}}$ & 0.53 & 60.95 & 6.55 & 0.70 \\
\textit{(F)} w/o $\mathcal{L}_{\text{content}}$ & 0.55 & 61.97 & 7.02 & 0.72  \\
\textit{(G)} w/o $\mathcal{L}_{\text{style}}$ & 0.54 & 62.34 & 6.78 & 0.73 \\
\noalign{\hrule height 1.2pt}
\end{tabular}
}
\label{tab:ablation}
\end{table}

\subsubsection{Study on SPPB}
As shown in Table~\ref{tab:ablation}, we ablate three aspects of the SPPB design.
Replacing the physics-grounded confidence map with a uniform weight ($\mathcal{W}\equiv0.5$, variant B) leads to consistent degradation across all metrics, confirming that fixed isotropic weighting fails to distinguish intrinsic from variation content.
Removing the Intrinsic-Variation disentanglement entirely (variant C) causes further drops, demonstrating that the $\mathcal{W}$-weighted projection is essential for suppressing modality-specific artifacts before fusion.
Freezing DINOv3 without LoRA adaptation (variant D) also hurts performance, indicating that task-aware fine-tuning is necessary for material-sensitive emissivity estimation.

\subsubsection{Study on RecA Loss}
We further ablate the components of $\mathcal{L}_{\text{RecA}}$ in Table~\ref{tab:ablation}.
Removing $\mathcal{L}_{\text{RecA}}$ entirely (variant E) degrades structural fidelity, confirming that DINOv3 feature-space alignment provides an essential semantic anchor during fusion training.
Ablating only $\mathcal{L}_{\text{content}}$ (variant F) causes semantic drift toward a single modality, while removing only $\mathcal{L}_{\text{style}}$ (variant G) introduces texture incoherence in the fused output.
These results validate that content and style alignment are complementary and jointly necessary for producing perceptually consistent fusion results.
\section{Conclusion}
\label{sec:conclusion}

We propose SGPDFuse, which replaces \emph{blind feature aggregation} with physics-disentangled representations via a Semantic-Physical Parametric Bridge, separating invariant scene attributes from transient environmental factors under the Intrinsic-Variation principle.
Building on this decomposition, the physics-guided fusion operates exclusively on intrinsic components, with Semantic Alignment via cosine similarity and Gram-matrix regularization providing self-consistent supervision in the DINOv3 feature space.
Extensive experiments demonstrate consistent state-of-the-art performance within a single unified architecture.

\bibliographystyle{ACM-Reference-Format}
\balance
\bibliography{main}


\end{document}